\documentclass{article}

\usepackage{microtype}
\usepackage{graphicx}
\usepackage{booktabs} 
\usepackage{amsfonts}
\usepackage{amsmath}
\usepackage{xcolor}
\usepackage{graphicx}
\usepackage{tikz}
\usepackage{wrapfig}
\usepackage{subcaption}
\usepackage{multirow}
\usepackage{tikz}
\usepackage[normalem]{ulem}

\usepackage{hyperref}

\usepackage[accepted]{icml2020}

\icmltitlerunning{On Variational Learning of Controllable Representations for Text without Supervision}

\newcommand\todo[1]{#1}

\begin{document}

\twocolumn[
\icmltitle{On Variational Learning of Controllable \\
  Representations for Text without Supervision}




\begin{icmlauthorlist}
\icmlauthor{Peng Xu}{to}
\icmlauthor{Jackie Chi Kit Cheung}{to,goo,ed}
\icmlauthor{Yanshuai Cao}{to}
\end{icmlauthorlist}

\icmlaffiliation{to}{Borealis AI}
\icmlaffiliation{goo}{McGill University}
\icmlaffiliation{ed}{Canada CIFAR Chair, Mila}

\icmlcorrespondingauthor{Peng Xu}{pxu4@ualberta.ca}

\icmlkeywords{Natural Language Processing, Variational Autoencoder, Controllable Text Generation}

\vskip 0.3in
]



\printAffiliationsAndNotice{}  

\begin{abstract}
The variational autoencoder (VAE) can learn the manifold of natural images on certain datasets, as evidenced by meaningful interpolation or extrapolation in the continuous latent space.
However, on discrete data such as text, it is unclear if unsupervised learning can discover a similar latent space that allows controllable manipulation.
In this work, we find that sequence VAEs trained on text fail to properly decode when the latent codes are manipulated, because the modified codes often land in holes or vacant regions in the aggregated posterior latent space, where the decoding network fails to generalize.
Both as a validation of the explanation and as a fix to the problem, we propose to constrain the posterior mean to a learned probability simplex, and perform manipulation within this simplex. 
Our proposed method mitigates the latent vacancy problem and achieves the first success in unsupervised learning of controllable representations for text.
Empirically, our method outperforms unsupervised baselines and strong supervised approaches on text style transfer, and is capable of performing more flexible fine-grained control over text generation than existing methods.
\end{abstract}

\section{Introduction}
\label{sec:intro}

High-dimensional data, such as images and text, are often generated through the interaction of many complex factors, such as lighting and pose in images or style and content in texts.
Recently, VAEs and other unsupervised generative models have found successes in modelling the manifold of natural images \citep{higgins2017beta,kumar2017variational,chen2016infogan}. These models often discover controllable latent factors that allow manipulation of the images through conditional generation from interpolated or extrapolated latent codes, often with impressive quality.
On the other hand, while various attributes of text such as sentiment and topic can be discovered in an unsupervised way, manipulating the text by changing these learned factors has not been possible with unsupervised generative models, to the best of our knowledge. \citet{cifka2018eval, kim2017adversarially} observed that text manipulation is generally more challenging compared to images, and the successes of these models cannot be directly transferred to texts.

Controllable text generation aims at generating realistic text with control over various attributes including sentiment, topic and other high-level properties.
The possibility of unsupervised controllable text generation could help in a wide range of applications such as dialogues systems \citep{wen2016network}.
Existing approaches \citep{shen2017style, fu2018style, li2018delete, sudhakar2019transforming} all rely on supervised learning from annotated attributes to generate the text in a controllable fashion.
The high cost of labelling large training corpora with attributes of interest limits the usage of these models, as pre-existing annotations often do not align with desired downstream goals.
Even if cheap labels are available, for example, review scores as a proxy for sentiment, the control is limited to the variation defined by the attributes.

In this work, we examine the obstacles that prevent sequence VAEs \cite{bowman2015generating} from performing well in unsupervised controllable text generation.
We empirically discover that manipulating the latent factors for typical semantic variations often leads to latent codes that reside in some low-density region of the aggregated posterior distribution. 
In other words, there are \emph{vacant} regions in the latent code space \citep{makhzani2015adversarial, rezende2018taming} not being considered by the decoding network, at least not at convergence.
As a result, the decoding network is unable to process such manipulated latent codes, yielding unpredictable generation results of low quality.
Although this issue has been raised in prior works, we provide direct evidence using topological data analysis to show that this vacancy problem is more severe for VAEs trained on text than image.

In order to mitigate the latent vacancy problem on text, we propose to constrain the posterior mean to a learned probability simplex and only perform manipulation within the probability simplex, which is referred as CP-VAE (Constrained Posterior VAE).
Two regularizers are added to the original objective of VAE.
The first enforces an orthogonal structure of the learned probability simplex; the other encourages this simplex to be filled without holes.
Besides confirming that latent vacancy is indeed a cause of failure in previous sequence VAEs', CP-VAE is also the first successful attempt towards unsupervised learning of controllable representations for text to the best of our knowledge.
Experimental results on text style transfer show that our method outperforms unsupervised baselines and strong supervised approaches, whose decoding network are trained from scratch.
Without supervision and the help of pre-training for generation, our method achieves comparable results with state-of-the-art supervised approaches leveraging large-scale pre-trained models for generation, with respect to the automatic evaluation metrics used in text style transfer.
Our proposed framework also enables finer-grained and more flexible control over text generation.
In particular, we can switch the topic in the middle of sentence generation, and the model will often still find a way to complete the sentence in a natural way, which has never been attempted by previous methods.\footnote{The code to reproduce our results can be found in \url{https://github.com/BorealisAI/CP-VAE}}




\section{Background: Variational Autoencoders}

The variational autoencoder (VAE) \citep{kingma2013auto} is a generative model defined by a prior $p(\pmb{z})$ and a conditional distribution $p_{\pmb\theta}(\pmb{x}|\pmb{z})$.
The VAE is trained to optimize a tractable variational lower bound of $\log p_{\pmb\theta}(\pmb{x})$:
\begin{equation}\label{eq:vae}
\begin{split}
\mathcal{L}_{\text{VAE}}(\pmb{x}; \pmb\theta, \pmb\phi) =& \mathbf{E}_{\pmb{z} \sim q_{\pmb{\phi}} (\pmb{z}|\pmb{x})}[\log p_{\pmb\theta}(\pmb{x}|\pmb{z})] \\
& - D_{\text{KL}}(q_{\pmb\phi}(\pmb{z}|\pmb{x})||p(\pmb{z})),
\end{split}
\end{equation}
where $q_{\pmb\phi}(\pmb{z}|\pmb{x})$ is a variational distribution parameterized by an encoding network with parameters $\pmb\phi$, and $p_{\pmb\theta}(\pmb{x}|\pmb{z})$ denotes the decoding network with parameters $\pmb\theta$.
This objective tries to minimize the reconstruction error to generate the data, and at the same time regularizes $q_{\pmb\phi}(\pmb{z}|\pmb{x})$ towards the prior $p(\pmb{z})$.
For text modelling, the input $\pmb x$ is some observed text. Both the encoding and decoding network are usually recurrent neural networks, and the model is called a sequence VAE.

Note that during learning, the decoding network $p_{\pmb\theta}(\pmb{x}|\pmb{z})$ only learns to decode $\pmb{z}$ that are sampled from $q_{\pmb\phi}(\pmb{z}|\pmb{x})$.
{\it In other words, the decoding network is never trained to decode the entire latent space.} Instead, it only learns to process $\pmb z$ sampled from the aggregated posterior distribution $q_{\pmb\phi}(\pmb z) = \mathbf{E}_{\pmb x \sim p_d(\pmb x)}q_{\pmb\phi}(\pmb z|\pmb x)$, where $p_d(\pmb x)$ is the training data distribution. If $q_{\pmb\phi}(\pmb z)$ has regions of low density, there is no guarantee that $p_{\pmb\theta}$ would generalize well to such places.
This is an important intuition that will become central to our analysis in Sec.~\ref{sec:moti}.

%
%
%

\section{Latent Vacancy Hypothesis}
\label{sec:moti}



\begin{table*}[h]
\footnotesize
\centering
\begin{tabular}{p{3cm} | p{4.5cm} | c | c | c} 
\hline
& Example & Transfer Strength & Content Preservation & NLL Discrepancy \\ \hline 
Source sentence & the pizza is offered without toppings and it 's lacking in flavor . & - & - & - \\ \hline
$\beta$-VAE w. aggr training ($\pm\sigma$) & the pizza is offered in toppings and it 's lacking in pittsburgh sauce . & Weak & Good & Small \\ \hline
$\beta$-VAE w. aggr training ($\pm2*\sigma$) & the pizza is more than fresh and your food is lacking in flavor & Medium & Medium & Medium \\ \hline
$\beta$-VAE w. aggr training (extremum) & the service is a great cut and the food is top notch in charlotte .& Strong & Bad & Large \\ \hline
CP-VAE (this work) & the pizza is full of spicy and it 's delicious . & Strong & Good & Small \\ \hline
\end{tabular}
\caption{Summary of the behaviours of $\beta$-VAE with aggressive training and our proposed CP-VAE. Detailed quantitative evaluations for transfer strength and content preservation are presented in Tab.~\ref{tab:unsupervised}.}
\label{tab:analysis}
\end{table*}

\begin{figure}[h]
\centering
 \includegraphics[width=.98\columnwidth]{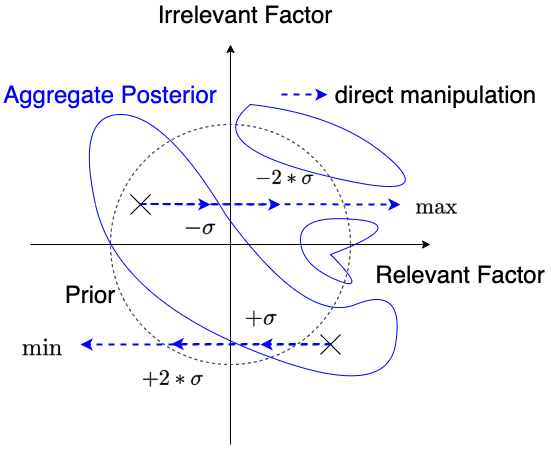}
 \captionof{figure}{Illustration of why latent vacancy prevents effective manipulation in VAEs. The aggregated posterior shown has multiple disconnected areas and direct manipulations of the relevant factor may fall into vacant regions of low density.} \label{fig:vae}
\end{figure}

We hypothesize that when trained on text data, the aggregated posterior of sequence-VAEs tend to have vacant regions of low density, where the decoder may fail to generalize to. The decoder could generalize to the vacant regions without ever seeing training examples, but there is no guarantee it can perform well in this case especially if the such vacancy is large. Fig.\ \ref{fig:vae} is an illustration of the intuition.

In this section, we conduct exploratory study on unsupervised sentiment manipulation and provide evidence from two different aspects to verify the above-mentioned hypothesis. First, we measure how the negative log-likelihood of latent codes under the aggregated posterior changes before and after manipulation. Second, since topology is the technical language to describe the notion of vacant regions or holes, we employ topological data analysis to confirm the exacerbation of latent vacancy problem on text as compared to images. In addition, we give a preview of our proposed method (later formally introduced in Section~\ref{sec:method}) and demonstrate that it avoids the latent vacancy problem using the same analyses.

\subsection{Unsupervised Sentiment Manipulation}


Here we describe the setup used to discover a sentiment latent dimension and subsequent exploration of manipulating the sentiment attribute.
Note that discovering sentiment feature in an unsupervised way is known to be possible, {\em e.g.}, in large-scale language models \citep{radford2017learning}.
However, limited success has been achieved for sequence VAE and its variants to change text attributes while preserving the relevant content, without annotated labels.

To perform unsupervised sentiment manipulation, we use the Yelp restaurant reviews dataset and the same data split following \citet{li2018delete}. 
We train a $\beta$-VAE \citep{higgins2017beta} with aggressive training of the encoder as proposed by \citet{he2019lagging}, which is the state of the art, and a significant improvement over vanilla sequence VAEs.
The model under study here has a latent space of $80$ dimensions with a LSTM encoder and decoder, with a $\beta$ of $0.35$. By inspecting the accuracy on the validation set, we find that there exists one dimension of latent code, $\pmb{z}_{[s]}$, achieving around $75\%$ sentiment classification accuracy by its value alone, while other latent codes get accuracy around $50\%$.
This means that this latent dimension is an effective sentiment indicator.
Further details can be found in Appendix~\ref{appendix:exploration:s1}-\ref{appendix:exploration:s2}.

However, when we try to perform sentiment manipulation by modifying this latent dimension, the decoding network fails to generate desirable outputs most of the time.
To ensure that the magnitude of manipulation suffices to change the sentiment of generated text, we try multiple magnitudes by moving $\pmb{z}_{[s]}$ (1) by $\sigma$; (2) by $2*\sigma$; (3) to $\min(\pmb{z}_{[s]})$ or $\max(\pmb{z}_{[s]})$, where $\sigma$, $\min$, $\max$ are the the standard deviation, the minimum and the maximum estimated on all the training samples. How we conduct the manipulation is illustrated in Fig.~\ref{fig:vae}.
We inspect the generated sentences with the manipulated codes to check whether they are transferred to the desired style successfully ({\em transfer strength}) and whether they are still relevant to the source sentence ({\em content preservation}).
We summarize the behaviours of $\beta$-VAE with aggressive training in Tab.~\ref{tab:analysis}, along with one randomly selected example for the purpose of illustration.
Although the sentiment can be flipped as we increase the magnitude of the manipulations, the transformed texts become irrelevant to the original text, meaning the content information in the latent code is ignored by the decoder.
On the other hand, when the manipulation on $\pmb{z}_{[s]}$ is small as in Fig.~\ref{fig:analysis} (A), $\beta$-VAE is unable to flip the sentiment of the transformed text, like the example in Tab.~\ref{tab:analysis}.
Detailed quantitative evaluations are presented in Sec.~\ref{sec:unsupervised}.

\subsection{NLL of the Codes under the Aggregated Posterior}

\begin{figure}
\centering
\includegraphics[width=.98\linewidth]{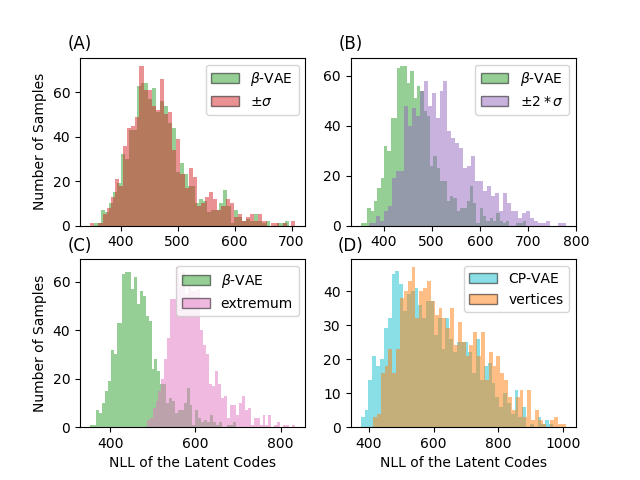}
\caption{Histograms of all the test samples' negative log-likelihood (NLL) under the aggregated posterior, considering their original latent codes and manipulated ones. (A) (B) (C): three manipulation strategies for $\beta$-VAE with aggressive training; (D) CP-VAE.}
\label{fig:analysis}
\end{figure}

To verify our hypothesis of vacant regions, we first compare the negative log-likelihood (NLL) of test samples' original latent codes as well as the manipulated ones, under the aggregated posterior.
An increase of the NLL after manipulation would indicate that the new codes land in regions of lower density. 
The aggregated posterior of our trained VAE is estimated with a large mixture of Gaussians where each component is the Gaussian posterior at one training data point.
Each test point's code (taken posterior mean) has an NLL under this mixture density. 
Fig.~\ref{fig:analysis} shows the histograms of NLLs of all $1000$ test samples' codes before and after manipulation.
We can see that the discrepancy in NLL between the original and the manipulated codes becomes larger as we increase the magnitude of the manipulation, indicating that the manipulated codes may fall into the low density area.

\subsection{Highest Density Region and Topological Analysis}
The notion of vacant regions or holes is a topological concept, so it is natural to use tools from topological data analysis (TDA) to measure and visualize this phenomenon. 
Given the aggregated posterior $q_{\pmb\phi}(\pmb z)$, the highest density region (HDR) at level $(1-\epsilon)$ \cite{hyndman1996computing} is defined to be:
$D_\epsilon = \{\pmb z | q_{\pmb\phi}(\pmb z) \geq c_\epsilon \}$, where $c_\epsilon$ is the largest constant such that $Pr(z \in D_\epsilon) \geq 1 - \epsilon$. Intuitively HDR captures the notion of ``significant support'', where we cut the density at $c_\epsilon$ to form a subset $D_\epsilon$ of the latent space that contains at least $1-\epsilon$ of the probability mass.  What we mean by the vacancy in the aggregated posterior $q_{\pmb\phi}(\pmb z)$ is that the $(1-\epsilon)$-HDR has holes or disconnected components. We want to emphasize that $\epsilon$ is conceptual and used to formalize the definition; it is not a hyperparameter of any model. In practice, whenever we draw a finite sample set, the points are in the HDR $D_\epsilon$ with probability $1 - \epsilon$, for some strictly positive $\epsilon$.

\begin{figure}
\centering
\includegraphics[width=.98\linewidth]{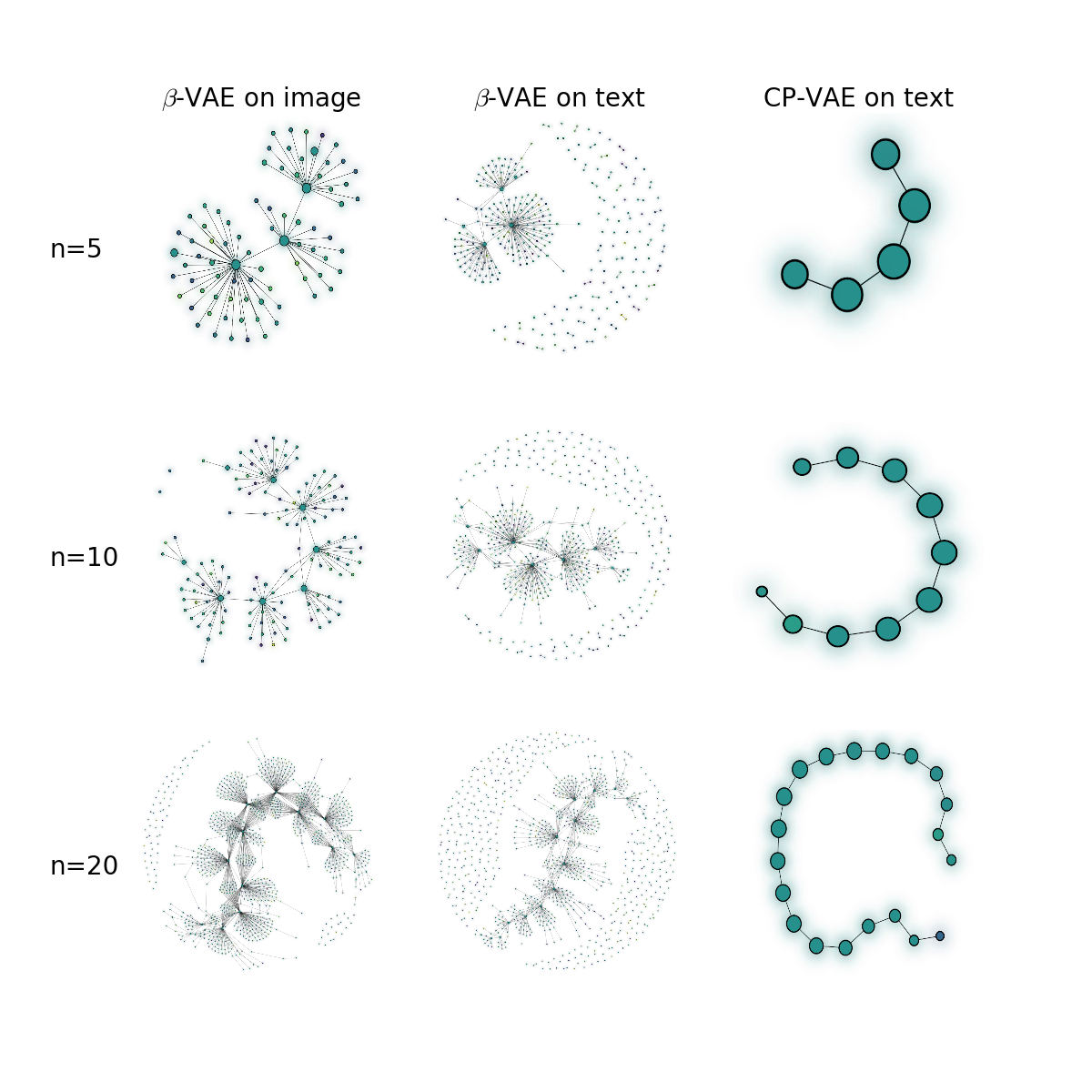}
\caption{Topological analysis of the highest density region (HDR) of aggregated posterior using the mapper algorithm. The connectedness of the graph holds the key topological information; the shape on the 2D plane is irrelevant. Different $n$'s control the coarseness of visualization. If a structure persists at multiple resolutions, it is stable. If it appears and disappears for selected value or a small range of $n$, then it is likely to be ``topological noise''. 
}
\label{fig:tda}
\end{figure}

We use the mapper algorithm \cite{singh2007topological} here to visualize the connectedness of $D_\epsilon$'s\footnote{In practice, we use the Kepler Mapper library by \citet{van2019kepler}} for $\beta$-VAE trained on images and text respectively.
Further details can be found in Appendix~\ref{appendix:exploration}.
The input to the mapper algorithm is a point cloud. For us, it is the posterior samples at training points under each model. The output of the mapper is a graph, like the ones in Figure~\ref{fig:tda}. Each node in the graph corresponds to a set of nearby points in the original point cloud. The connectivity of the graph reflects some topological properties of the sampling space of the point cloud. Such properties include connectedness and the presence of holes.

The main take-away, as shown in Fig.~\ref{fig:tda}, is that the HDR of $\beta$-VAE on images is one connected component (up to topological noise on the finest scale); whereas, for text, there are many disconnected components across all scales of visualization. This observation suggests that the underlying $D_\epsilon$ for $\beta$-VAE on text is disconnected, providing empirical evidence that the latent vacancy problem is more severe on text than on images.
Further explanations about the relationship of connectedness of $D_\epsilon$ and that of the mapper graphs can be found in Appendix \ref{appendix:tda}.





\subsection{Constraining the posterior}

In order to resolve the latent vacancy problem, we propose CP-VAE in this work, where we constrain the posterior in such a way that the manipulation only happens in a learned simplex, so that most space in this constrained subspace can be covered during training.
In this constrained subspace, the phenomenon of low density holes of aggregated posterior is significantly reduced, as Fig.~\ref{fig:analysis} (D) empirically show that there is little change in NLL of original versus manipulated codes. Furthermore, Fig.~\ref{fig:tda} shows that the HDR of CP-CAE is one connected component\footnote{The HDR visualized here is for $\pmb{z}^{(1)}$ introduced in Sec.~\ref{sec:method}}.
At the same time, CP-VAE can maintain its transfer strength to effectively transfer the source sentence to the desired style, as exemplified in Tab.~\ref{tab:analysis}.
The details of our method are presented in the next section.

\section{Method}
\label{sec:method}

\subsection{Overview}

The experiments conducted in Sec.~\ref{sec:moti} validate the existence of vacancy in the aggregated posterior latent space.
One potential way to resolve the problem is to better match the aggregated posterior with the prior \citep{makhzani2015adversarial,tomczak2017vae,kim2017adversarially}. However, in terms of unsupervised learning of controllable representation for text, these previous methods have not shown success; \citet{kim2017adversarially} only attempted supervised text style transfer, and also reported negative results from the AAE \citep{makhzani2015adversarial}. 
Another way to resolve the vacancy issue is to directly enforce that the aggregated posterior itself has no vacant region anywhere where we would like to perform latent code manipulation.
We propose to map the posterior Gaussian mean to a constrained space, more specifically a learned probability simplex, where we can encourage the constrained latent space to be filled without vacancy, and perform manipulation to be within this simplex. 
We add a mapping function as part of the encoding network which maps the mean of the Gaussian posterior to a constrained space.
Two regularization terms are introduced to ensure the learned simplex is not degenerate and that this subspace is well filled. 

In addition, we model the relevant factors that we wish to control separated from the irrelevant factors by splitting $\pmb{z}$ into two parts, $\pmb{z}^{(1)}$ and $\pmb{z}^{(2)}$, following prior work \citep{bao2019generating}.
The first part captures the relevant factors that are dominant in the data without an inductive bias from external signals, while the second part learns to encode the remaining local information that is useful for reconstructing the source sentences. 
As a result, $q_{\pmb\phi}(\pmb{z}|\pmb{x})$ is decomposed into $q_{\pmb\phi_1}(\pmb{z}^{(1)}|\pmb{x})q_{\pmb\phi_2}(\pmb{z}^{(2)}|\pmb{x})$ where $\pmb\phi = \pmb\phi_1 \cup \pmb\phi_2$. With diagonal covariances, the KL divergence term in Eq.\ \ref{eq:vae} splits into two separate KL terms.
In practice, we use a MLP encoding network to parametrize $\pmb z^{(1)}$ with some sentence representation as the input ({\em e.g.}, averaging GloVe embeddings \citep{pennington2014glove} over the input tokens) and a LSTM encoding network to parametrize $\pmb z^{(2)}$.
We only constrain the posterior of $\pmb z^{(1)}$, and $\pmb z^{(2)}$ is optimized the same way as the traditional VAE.

\subsection{Constraining the Posterior}

We now describe how to map the mean $\pmb\mu$ of the Gaussian posterior for $\pmb{z}^{(1)}\in\mathbb{R}^N$ to a constrained latent space.
We would like to constrain the mean $\pmb\mu$ to have a structure as follows:
\begin{equation} \label{eq:structure}
\pmb\mu = \sum_{i=1}^Kp_i\pmb{e}_i,  \sum_{i=1}^Kp_i =1,  \langle \pmb{e}_i, \pmb{e}_j \rangle=0, i \neq j,  K \le N
\end{equation}
where $\pmb{e}_i$ are vectors representing the relevant factors, $p_i$ is the proportion of $i$th relevant factor encoded in $\pmb{z}^{(1)}$ and $K$ is a hyperparameter indicating the number of relevant factors to discover.
In other words, the mean of the Gaussian posterior of $\pmb{z}^{(1)}$ is constrained to be inside a $K$-dimension probability simplex in $\mathbb{R}^N$ whose vertices are represented by the orthogonal basis vectors $\pmb{e}_i, i=1,\dots,K$.
Given the outputs of the MLP encoder $\pmb h$ and $\log \pmb\sigma^2$, we learn an additional mapping function $\pi$ which maps $\pmb h$ to the constrained posterior space, which can be treated as part of the encoding network:
\begin{equation}
\pmb\mu = \pi(\pmb{h}) = \pmb E \cdot \text{softmax}(\pmb{W} \pmb{h} + \pmb{b}),
\end{equation}
where $\pmb{E} = [\pmb{e}_1, \dots, \pmb{e}_K]$ is a learnable embedding matrix representing the bases, $\pmb{W}$ is the learnable weight matrix, and $\pmb{b}$ is the learnable bias vector.
As a result, the constrained posterior is parametrized by $\pmb\mu$ and $\log\pmb\sigma^2$ as a Gaussian distribution $\mathcal{N}(\pmb{\mu}, \text{diag}(\pmb\sigma^2))$.

With the mapping function alone, the proposed VAE suffers from posterior collapse \citep{bowman2015generating}, a well-known problem where the model ignores the latent code $\pmb{z}$ during the training. 
Further complicating matters is the fact that there is an abundance of signals for predicting the next token in the text, but the signals indicating high-level semantics are quite sparse.
It is thus unlikely that the VAEs can capture useful relevant factors from raw text without collapse.
For these reasons, we enforce orthogonality in the learnt basis vectors as defined in Eq.~\ref{eq:structure}, which introduces a natural recipe to prevent posterior collapse for $\pmb z^{(1)}$.
Note that the KL divergence between $q_{\pmb\phi_1}(\pmb{z}^{(1)}|\pmb{x})$ and $p(\pmb{z}^{(1)})$ is
\begin{equation}
\begin{split}
D_{\text{KL}}&(q_{\pmb\phi_1}(\pmb{z}^{(1)}|\pmb{x})\Vert p(\pmb{z}^{(1)})) = \\ 
& \frac12 \pmb\mu^\top \pmb\mu +\frac12 \left( \pmb\sigma^\top\pmb\sigma - \log\pmb\sigma^\top\pmb\sigma - 1 \right).
\end{split}
\end{equation}
With orthogonality in the basis vectors, the first term in the above equation can be factorized into
\begin{equation}\label{eq:factor}
\pmb\mu^\top\pmb\mu = (\sum_i p_i \pmb{e}_i)^\top(\sum_i p_i \pmb{e}_i) = \sum_i p_i^2 \pmb{e}_i^\top\pmb{e}_i.
\end{equation}
To encourage orthogonality in the basis vectors, a regularization term is added to the objective function:
\begin{equation}\label{eq:reg}
\mathcal{L}_{\text{REG}}(\pmb{x};\pmb\phi_1)=\lVert \pmb{E}^\top \pmb{E}  - \alpha\pmb{I}\rVert,
\end{equation}
where $\pmb{I}$ is the identity matrix and $\alpha$ is a hyperparamter.
When $\mathcal{L}_{\text{REG}}=0$, $\pmb{e}_i^\top\pmb{e}_i=\alpha$.
In this case, $\pmb\mu^\top\pmb\mu=\alpha\sum_i p_i^2$ reaches its minimum $\frac\alpha{K}$ when $\pmb{p}$ is a uniform distribution. 
The proof can be found in Appendix~\ref{appendix:proof}.
In practice, $\mathcal{L}_{\text{REG}}$ will quickly decrease to around $0$, ensuring that the KL term will never fully collapse with the structural constraint.
When it comes to controlled generation, one can choose a vertex or any desired point in the probability simplex.

\subsection{Filling the Constrained Space}

Constraining the posterior inside a certain space does not guarantee that this space will be filled after training.
We also need to encourage the probability distribution over the relevant factors $\pmb p$ to cover as much of the constrained latent space as possible.
We introduce a reconstruction error of the structured latent code in order to push $\pmb p$ away from a uniform distribution.
For each input sentence, we randomly sample $m$ sentences from the training data as negative samples.
By applying the same encoding process, we get the structured latent code $\pmb{\mu}^{(-)}_i$ for each negative sample.
Our goal is to make the raw latent code $\pmb{h}$ similar to the restructured latent code $\pmb{\mu}$ while different from latent codes $\pmb{\mu}^{(-)}_i$ of the negative samples, so that $\pmb p$ is generally different for each input sample.
The structured reconstruction loss is formulated as a margin loss as follows:
\begin{equation} \label{eq:s-rec}
\begin{split}
&\mathcal{L}_{\text{S-REC}}(\pmb{x};\pmb\phi_1) =\\
&\mathbb{E}_{\pmb{z}^{(1)}\sim q_{\pmb\phi_1}(\pmb{z}^{(1)}|\pmb{x})} \left[\frac1m\sum_{i=1}^m \max(0, 1 - \pmb{h} \cdot \pmb{\mu} + \pmb{h} \cdot \pmb{\mu}^{(-)}_i)\right].
\end{split}
\end{equation}
Our final objective function is defined as follows:
\begin{equation}\label{eq:loss}
\mathcal{L}(\pmb{x}; \pmb\theta, \pmb\phi) = \mathcal{L}_{\text{VAE}} + \mathcal{L}_{\text{REG}} + \mathcal{L}_{\text{S-REC}}.
\end{equation}

\section{Experiments}
\label{sec:exp}

To demonstrate the effectiveness of CP-VAE, we compare it to unsupervised baselines with $\beta$-VAE and state-of-the-art optimizing techniques, considering the performance on unsupervised sentiment manipulation.
Following evaluation protocols in text style transfer, we also compare our method to strong supervised approaches.
Furthermore, we showcase the ability of finer-grained style discovery and transition possessed by our system, which has not been attempted in the literature. 
Detailed configurations including the hyperparameters, model architecture, training regimes, and decoding strategy are found in Appendix~\ref{appendix:experiments}.

\subsection{Comparisons with Unsupervised Baselines}
\label{sec:unsupervised}

\begin{table}[h]
\centering
\scriptsize
\caption{Comparisons with unsupervised baselines on Yelp dataset.}
\begin{tabular}{l | l  l }
\hline
Model & Accuracy (AC) $\uparrow$ & BLEU (BL) $\uparrow$ \\ \hline
$\beta$-VAE ($\pm\sigma$) & $50.98 \pm 2.89 $ & $4.02 \pm 0.77$\\
$\beta$-VAE ($\pm2*\sigma$) & $78.44 \pm 4.84$ & $1.49 \pm 0.29$\\
$\beta$-VAE (extremum) & $98.18 \pm 1.56 $ & $0.56 \pm 0.40$\\ \hline
$\beta$-VAE w. aggr training ($\pm\sigma$) & $26.76 \pm 6.44$ & $27.91 \pm 4.39$ \\
$\beta$-VAE w. aggr training ($\pm2*\sigma$) & $57.46 \pm 14.47 $ & $11.73 \pm 6.74$ \\
$\beta$-VAE w. aggr training (extremum)&  $88.08 \pm 14.95 $ & $4.57\pm4.63$ \\ \hline
CP-VAE w. GloVe & $60.22 \pm 4.57$ &  $33.69 \pm 1.47$  \\
\hspace{1em} without $\mathcal{L}_{\text{REG}}$ & $10.82 \pm 0.91$ & $33.27 \pm 2.84$ \\ 
\hspace{1em} without $\mathcal{L}_{\text{S-REC}}$ & $12.28 \pm 3.69$ & $49.34 \pm 2.65$\\
\hline
\end{tabular}
\label{tab:unsupervised}
\end{table}

\begin{figure}[ht]
\begin{center}
 \includegraphics[width=0.9\columnwidth]{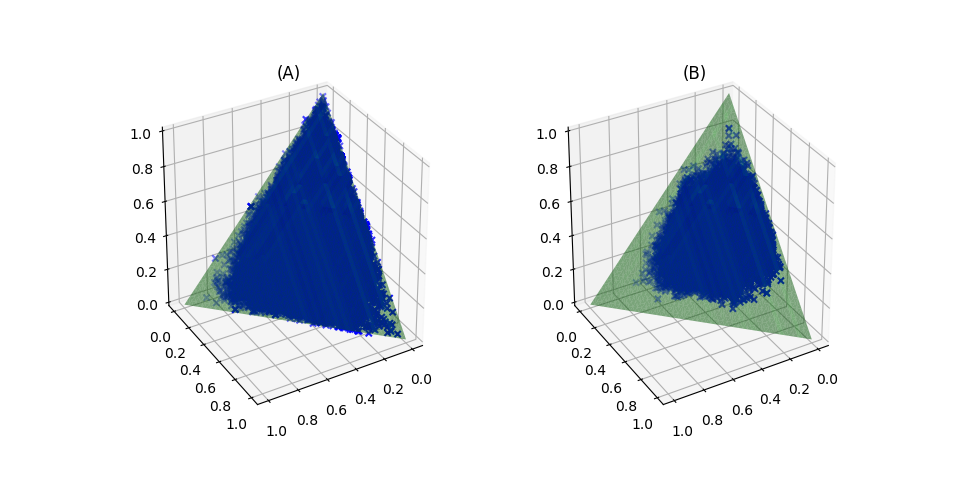}
\end{center}
\caption{\small Visualization of all training samples in the probability simplex: (A) With $\mathcal{L}_{\text{S-REC}}$ ;(B) Without $\mathcal{L}_{\text{S-REC}}$.}
\label{fig:simplex}
\end{figure}

\vspace{-1em}

\paragraph{Experimental setup:}
We use the same experimental setting and dataset as mentioned in Sec.~\ref{sec:moti}.
The $80$D latent code is split into $16$ and $64$ dimensions for $\pmb{z}^{(1)}$ and $\pmb{z}^{(2)}$ respectively.
The sentence representations for $\pmb{z}^{(1)}$ is the averaged GloVe embeddings over the input tokens and $K$ is chosen as $3$.
To decide which basis vector corresponds to which sentiment, we sample $10$ positive and $10$ negative sentences in the development set, pass them to the encoder, and choose the basis vector with the highest average $p_i$ in $\pmb p=\text{softmax}(\pmb W \pmb{h} + \pmb b)$, yielding $v_p$ as the positive basis and $v_n$ as the negative basis.
If $v_p$ and $v_n$ are chosen to be the same vector, we choose the index with the second highest $p_i$ for $v_p$.
To perform sentiment manipulation, we fix $\pmb{z}^{(1)}$ to be the chosen basis vector; that is, $v_p$ or $v_n$.


\newcommand{\xmark}{$\mathbin{\tikz [x=1.4ex,y=1.4ex,line width=.2ex] \draw (0,0) -- (1,1) (0,1) -- (1,0);}$}
\newcommand{\cmark}{$\checkmark$}
\begin{table*}[h]
    \centering
   \footnotesize
\caption{\small Comparisons with supervised approaches on Yelp and Amazon dataset.}
\begin{tabular}{l c c l l l l c l l l l}
\hline
& & & \multicolumn{4}{c}{Yelp} && \multicolumn{4}{c}{Amazon} \\ \cline{4-7} \cline{9-12}
Model &Supervised&GPT-2& AC $\uparrow$ & BL $\uparrow$ & GL $\uparrow$ & PL $\downarrow$&& AC $\uparrow$& BL $\uparrow$& GL $\uparrow$& PL $\downarrow$\\ \hline
Source &-&-& 1.8 & 100.0 & 8.4 & 26.6 && 16.3 & 100.0 & 22.8 & 34.5 \\
Human &-&-& 70.1 & 25.3 & 100.0 & 63.7 && 41.2 & 45.7 & 100.0 & 68.6 \\ \hline
CA &\cmark&\xmark& 74.0 & 20.7 & 6.0 & 103.6 && {\bf 75.5} & 0.0 & 0.0 & {\bf 39.3}\\
SE &\cmark&\xmark& 8.2 & {\bf 67.4} & 6.9 & 65.4 && 40.2 & 0.4 & 0.0 & 125.0 \\
MD &\cmark&\xmark& 49.5 & 40.1 & 6.6 & 164.1 && 70.1 & 0.3 & 0.0 & 138.8 \\
D\&R &\cmark&\xmark& {\bf 88.1} & 36.7 & 7.9 & 85.5 && 49.2 & 0.6 & 0.0 & 46.3 \\ \hline
{\bf CP-G} &\xmark&\xmark& 66.7 & 35.5 & 7.5 & 67.8 && 60.1 & {\bf 35.4} & {\bf 11.5} & 109.1 \\
{\bf CP-B} &\xmark&\xmark& 55.4 & 48.4 & {\bf 9.6} & {\bf 47.6} && 40.0 & {\bf 39.7} & {\bf 12.7} & 97.3\\ \hline \hline
B-GST &\cmark&\cmark& 85.6 & 45.2 & 12.7 & 49.6 && 55.2 & 52.3 & 18.1 & 48.2 \\
\hline
\end{tabular} 
\label{tab:supervised}
\end{table*}

\begin{table*}[h]
\centering
\footnotesize
\caption{\small Samples of generated sentences. SRC is the input sentence.}
\begin{tabular}{| l | p{5.5cm} | p{5.5cm} |}
\hline
{\bf Yelp} & {\itshape Positive to Negative} & {\itshape Negative to Positive} \\ \hline
SRC & this place is super yummy !& but it probably sucks too !\\ \hline
B-GST & this place is super bad !& but it tastes great too ! \\ \hline
{\bf CP-G} & this place is super slow and watered down .& but it 's truly fun and insanely delicious .\\ \hline
{\bf CP-B} & this place is super greasy and gross !& but it 's probably wonderful when you ! \\ \hline
\hline
{\bf Amazon} & {\itshape Positive to Negative} & {\itshape Negative to Positive} \\ \hline
SRC & because it s made of cast iron , scorching is minimized . & they are cheerios, afterall, and we love the original kind .\\ \hline
B-GST & because it s cheaply made of cast iron , is useless .& they are sturdy, afterall, sturdy and we love the original .\\ \hline
{\bf CP-G} & because it s made of cast iron , vomitting .& they are ripe, tastier , and we love them .\\ \hline
{\bf CP-B} & because it s made of cast iron , limp .& they are divine, fluffier , and we love them .\\ \hline
\end{tabular} 
\label{tab:samples}
\end{table*}

\paragraph{Comparsions with metrics on text style transfer:}
For quantitative evaluation, we adopt two general automatic evaluation metrics used in text style transfer \citep{fu2018style,li2018delete,sudhakar2019transforming}: classification accuracy (AC) of a pre-trained classifier to measure the transfer strength; BLEU score (BL) of the transferred sentences against the source sentences to measure the content preservation.
As shown in Tab.~\ref{tab:unsupervised}, $\beta$-VAE alone performs poorly in terms of content preservation no matter the modification magnitude, while aggressively training the encoder can notably help improve content preservation.
However, no matter we use aggressive training or not, the content preservation deteriorates drastically as we increase the modification magnitude, in order to achieve reasonable transfer strength.
With large enough modification magnitude, the classification accuracy can be pushed to almost perfect, while the BLEU score decreases towards zero, meaning that the transferred sentences become totally irrelevant to the source sentences.
The results match our observations from the experiments on density under the aggregated posterior distribution, confirming that latent vacancy prevents effective manipulation of the latent codes.
To the contrary, CP-VAE can achieve much better content preservation while maintain its transfer strength, indicating its effectiveness to mitigate the latent vacancy problem.

\paragraph{Ablation study:}
We also conduct an ablation study by removing $\mathcal{L}_{\text{REG}}$ and $\mathcal{L}_{\text{S-REC}}$ from the objective.
The results demonstrate that both terms are crucial to the success of CP-VAE.
Without $\mathcal{L}_{\text{REG}}$, CP-VAE experiences posterior collapse for $\pmb z^{(1)}$.
As a result, $v_p$ and $v_n$ collide with each other, leading to failure in disentangled representation learning.
Since we choose $K$ as $3$, it is convenient to visualize the samples during training with $\pmb p$ in the learnt probability simplex, as shown in Fig.~\ref{fig:simplex}.
We can see that the whole simplex is mostly covered with samples with the help of $\mathcal{L}_{\text{S-REC}}$.
Without $\mathcal{L}_{\text{S-REC}}$, the decoding network fails to recognize the basis vectors due to the poor coverage of the probability simplex, causing the model to lose most of its transfer strength.

\todo{
At the same time, we do not claim that there are no other necessary conditions for the success of CP-VAE.
First, if $\pmb z^{(1)}$ uses raw text as inputs with a LSTM encoder, the VAEs will ignore $\pmb{z}^{(1)}$ by making all the samples collapse to one vertex on the simplex.
On the other hand, if $\pmb z^{(2)}$ uses pre-trained embeddings with pooling like $\pmb z^{(1)}$ as inputs, the VAEs would be unable to reconstruct the source sentence effectively, because the representations would lose most local information necessary for the reconstruction.
However, this necessity is beside the point of our paper and does not contradict the evidence we presented for the latent vacancy hypothesis.
}

\begin{table*}[h]
\centering
\footnotesize
\caption{\footnotesize Two pairs of samples generated without and with topic transition. The first sentence in the pair is generated with a topic fixed throughout the generation; while the second sentence is generated with topic transition, the generated outputs after switching are marked as bold.}
\begin{tabular}{l | p{10.2cm}}
\hline
{\em World} throughout & A federal judge on Friday ordered a federal appeals court to overturn a federal appeals court ruling that the Visa and MasterCard credit card associations violated federal antitrust law by barring the names of the state .\\ \hline
{\em World} to {\em Sci/Tech} & A federal judge on Friday ordered a federal appeals court to overturn a decision by the Supreme Court to {\bf overturn a decision by the Federal Communications Commission to block the company's antitrust case against Microsoft Corp .}\\ \hline \hline
{\em Sports} throughout & NEW YORK (Reuters) - Roger Federer, the world's No. 1 player, will miss the rest of the season because of a sore quadriceps . \\ \hline
{\em Sports} to {\em Business} & NEW YORK (Reuters) - Roger Federer, the world's No. 1 player, will miss the rest of the {\bf year because of a bid-rigging scandal .} \\ 
\hline
\end{tabular}
\label{tab:transition}
\end{table*}

\subsection{Comparisons to Supervised Approaches on Text Style Transfer}

\paragraph{Experimental setup:}
We choose two datasets, Yelp and Amazon, used in works \citep{li2018delete, sudhakar2019transforming} on text style transfer which provide human gold-standard references for the test set.
The same train-dev-test splits are used in our experiments.
Two different sentence representations are used in this experiment, averaged GloVe and BERT \citep{devlin2018bert}, denoted as {\bf CP-G(loVe)} and {\bf CP-B(ert)} respectively.
The remaining settings are as described in the above section.

\paragraph{Compared supervised approaches:}
On the two datasets, we compare to three adversarially trained models: StyleEmbedding ({\bf SE}) \citep{fu2018style}, MultiDecoder ({\bf MD}) \citep{fu2018style}, CrossAligned ({\bf CA}) \citep{shen2017style} and two state-of-the-art models based on a ``delete, transform, and generate'' framework: DeleteAndRetrieve ({\bf D\&R}) \citep{li2018delete} and Blind-GenerativeStyleTransformer ({\bf B-GST}) \citep{sudhakar2019transforming}.
To be noted, the decoding network of {\bf B-GST} is based on GPT-2 \citep{radford2019language}, while all the other models including ours train the decoding network from scratch.

\paragraph{Evaluation protocols:}
Four different automatic evaluation metrics are used to measure the different perspectives of the transferring quality, following \citet{sudhakar2019transforming}.
To measure transfer strength, we use pre-trained CNN based classifiers achieving 98\% and 84\% accuracies on the test sets of Yelp and Amazon respectively. 
To measure content preservation, we use the BLEU \citep{papineni2002bleu} score of the transferred sentences against the source sentences.
To measure fluency, we finetune OpenAI GPT-2 \citep{radford2019language} with 345 million parameters on the same training-dev-test split to obtain the perplexity of generated sentences.
The fine-tuned language models achieve perplexities of $26.6$ and $34.5$ on the test sets of Yelp and Amazon respectively.
In addition, \citet{sudhakar2019transforming} argued that the Generalized Language Evaluation Understanding Metric (GLEU) has a better correlation with the human judgement.
Here, we use the implementation of GLEU\footnote{https://github.com/cnap/gec-ranking} provided by \citet{napoles2015ground} to calculate the GLEU score.

\paragraph{Result Analysis:}
As observed by \citet{li2018delete} and \citet{sudhakar2019transforming}, accuracy, BLEU score and perplexity do not correlate well with human evaluations.
Therefore, it is important to not consider them in isolation.
Tab.~\ref{tab:supervised} shows that our proposed approaches get similar scores on these metrics with human reference sentences on the second row, indicating that the generated sentences of our proposed approaches is reasonable considering the combination of these metrics.
As seen by \citet{sudhakar2019transforming} and verified in Sec.~\ref{sec:unsupervised}, GLEU strike a balance between target style match and content retention and correlate well with the human evaluations.
From Tab.~\ref{tab:supervised}, CP-VAE consistently outperforms the three adversarially trained models and D\&R on GLEU by a noticeable margin.
As compared to B-GST, the current state-of-the-art, which leverages GPT-2 for generation, the results are still competitive, despite the fact that CP-VAE is trained unsupervisedly and from scratch.
By checking the samples generated from the models as shown in Tab.~\ref{tab:samples}, B-GST is more consistent to the source sentence, which can be expected, since it only makes necessary edits to flip the sentiment.
CP-VAE tends to generate more diverse contents which may not be relevant sometimes, but the overall quality is reasonable.
More samples can be found in Appendix~\ref{appendix:transfer}.

\subsection{Finer-grained Style Discovery and Transition}

To further explore the potential of CP-VAE, we conduct the following exploratory experiments.
We use the AG news dataset constructed by \citep{zhang2015character}, which contains four topic categories which are {\em World, Sports, Business} and {\em Sci/Tech}, with the title and description fields.
Here, we drop the title and just use the description field to train CP-VAE and set $K=10$.
All four topics are automatically discovered by CP-VAE and identified as described in Sec.~\ref{sec:unsupervised}.
We also compare the results of our identified topics to standard baselines for unsupervised topic modelling, the details can be found in Appendix~\ref{appendix:topic}.
We choose a basis vector discovered by our model and generate a few tokens.
Then, we switch the basis vector and continue the generation until the {\em end-of-seq} token is generated.
Generated samples are shown in Table~\ref{tab:transition}.
We see that our model learns to transition from one topic to another in a natural and fluent way within the same sentence.
Several observations can be made based on these samples: (1) it is good at detecting name entities and replacing them with the name entities related to the chosen topic; (2) there is no hard restriction on when to switch the topic; the model will determine an appropriate way to do the transition by itself. 
Such observations confirm that CP-VAE possesses a filled constrained latent space which make the latent code robust to manipulation across different time steps, which can be effectively reflected in the generation process.
Due to space limitations, we put more samples in Appendix~\ref{appendix:transition}.

%
%
%

\section{Related Work}

\subsection{Unsupervised Learning of Disentangled Representations}

Learning disentangled representations is an important step towards better representation learning \citep{bengio2013representation} which can be useful for (semi-)supervised learning of downstream tasks, transfer and few-shot learning \citep{peters2017elements}.
VAEs have achieved promising results for unsupervised learning of disentangled representations.
Several variations of VAEs have been proposed for better disentanglement \citep{higgins2017beta, kumar2017variational, chen2016infogan, razavi2019preventing}.
However, progress in this direction has been restricted to the image domain, and does not demonstrate successful controlled generation on text.

\subsection{Controlled Text Generation}

In order to perform controllable text generation, previous methods either assume annotated attributes or multiple text datasets with different known styles \citep{hu2017toward, shen2017style,kim2017adversarially,fu2018style,li2018delete, sudhakar2019transforming, logeswaran2018content, lample2018multiple}.
The requirement of labelled data largely restricts the capabilities and the applications of these models. 
Instead, all our proposed framework needs is raw text without any annotated attribute.

\section{Conclusion}
In this work, we investigate latent vacancy as an important problem in unsupervised learning of controllable representations when modelling text with VAEs.
To mitigate this, we propose to constrain the posterior within a learned probability simplex and encourage this space to be filled, achieving the first success towards controlled text generation without supervision.
However, the constrained posterior also means that the aggregated posterior can never match the isotropic Gaussian prior which points to a potential future direction to resolve this mismatch by selecting or learning a better prior as in \citep{tomczak2017vae}.

\section*{Acknowledgements}

Thanks to Ivan Kobyzev for the useful discussion and feedback, and to all the anonymous reviewers for their valuable inputs.

\bibliography{ref}
\bibliographystyle{icml2020}
\newpage
\appendix

\onecolumn
\section{Experimental Details for Reproducibility}

All the codes in this paper are implemented with PyTorch.
For the implementation of $\beta$-VAE and pre-processing step in this paper, we follow the codebase of \cite{he2019lagging}: \url{https://github.com/jxhe/vae-lagging-encoder}.
The datasets used in Sec.~\ref{sec:exp} can be found in the codebase of \citet{sudhakar2019transforming}: \url{https://github.com/agaralabs/transformer-drg-style-transfer}.

\section{Details about Exploratory Experiments}
\label{appendix:exploration}

\subsection{Model Details for Unsupervised Sentiment Manipulation}
\label{appendix:exploration:s1}

For the $\beta$-VAE we used for the unsupervised sentiment manipulation, we use a LSTM encoding network and a LSTM decoding network.
For the encoding network, the input size is 256, and the hidden size is 1,024.
For the decoding network, the input size is 256, the hidden size is 1,024, and dropouts with probability 0.5 are applied on after the embedding layer and the LSTM layer in the decoding network.
The dimension for the latent code is 80, and the batch size is 32.
We use SGD with learning rate 1.0 to update the parameters for both the encoding and the decoding network.
We train the model until the reconstruction loss stops decreasing.
For aggressive training, we follow \citet{he2019lagging} to aggressively train the encoding network.
Those hyperparameters are chosen following the experiments conducted in Sec.~\ref{sec:exp} without extra tuning.

In \citet{higgins2017beta}, it is encouraged to set $\beta >$ to achieve better disentangling performance than vanilla VAEs. 
However, a large $\beta$ will push the reconstruction loss higher, and the KL loss lower, leading to terrible content preservation from the source sentence.
In practice, with a large $\beta$, the classification accuracy can be easily pushed to be perfect by always generating pivot sentences like "Great!" and "Terrible!".
In order to achieve the best trade-off between transfer strength and content preservation, we chose $\beta$ as 0.35 by inspecting the generated sentences.

\subsection{Identifying the Latent Factor Indicating the Sentiment}
\label{appendix:exploration:s2}

First, we normalize the value of each latent code by subtracting the mean estimated over all the training samples.
Then we use the polarity of each latent code to classify the sentiment in the validation set.
The one with the highest accuracy is identified as the latent factor indicating the sentiment.

\subsection{Details for $\beta$-VAE Trained on Images}
\label{appendix:exploration:image}

For the $\beta$-VAE trained on OMNIGLOT, we use the exact same setting following the codebase of \cite{he2019lagging}: \url{https://github.com/jxhe/vae-lagging-encoder}.

\subsection{Topological Analysis: Connectedness of Mapper Graph}
\label{appendix:tda}
To help interpret the visualization in Figure~\ref{fig:tda}, we give a brief description of the mapper algorithm. For a more technical introduction and details, please see \citet{singh2007topological}. 
The algorithm requires some user-specified options. The first one is a continuous function $f$ (also called a ``lens'', filter or projection) that points $\pmb z$ from a point cloud $Z$ to $\mathbb{R}$. The range of $f$, $I = f(Z)$, is then divided into $n$ overlapping open intervals $\{I_{j}\}^n_{j=1}$. We then find the pre-images of these intervals, $U_j = f^{-1}(I_j)$, which are open sets in the input space. Points in each $U_j$ are then further partitioned using a clustering algorithm (e.g. DBSCAN). In the end, across all pre-images $U_j$'s, we have a collection of clusters $U_{jk}$, which might or might not intersect. We represent each cluster $U_{jk}$ as a graph node and connect two nodes if and only if the point sets intersect. Here we take the continuous function to be the sum of values in each dimension of input, and we vary $n$ to inspect if the discovered structure persists over multiple scales or is a noise. 

In the resulting graph, disconnected nodes can arise in two ways. First, if the intersecting portion of some pair of overlapping intervals does not actually contain a point mapped from the input point cloud. But this is avoided by the open cover construction in the implementation of \citet{van2019kepler}. The second case is if there are actually disconnected components in the input space. Without loss of generality, assume there are two. Then by construction, some points from the two sets will be mapped to the same interval $\tilde{I}$, or shared portion of two covering intervals in the range, $\tilde{I} = I_{l} \cap I_k $. The pre-image of $\tilde{I}$ is the only set that could lead to a connection of the nodes, however, since it contains points that are not in the same neighborhood, clustering of this pre-image will produce two disconnected nodes, forming a disconnected graph.

\subsection{Details for Topological Data Analysis}
For the mapper algorithm, we use DBSCAN as the clustering algorithm.
For DBSCAN, we set $\epsilon=0.1$ and $\text{min\_samples}=3$.
We sample $100,000$ points from the training set as the input.
For the three cases we visualize, the latent dimensions are all $16$.
We choose the first $16$ dimension for $\beta$-VAE trained on text and images.
For CP-VAE, we use $\pmb{z}^{(1)}$.

%
%
%

\section{Details about Experiments on Text Style Transfer}
\label{appendix:experiments}
\subsection{Training Regimes}

Across all the datasets, we use Adam with learning rate 0.001 to update the parameters for the encoding network, while SGD with learning rate 1.0 to update the parameters for the decoding network.
The batch size is chosen to be 32.
Dropouts with drop probability 0.5 are applied on applied on after the embedding layer and the LSTM layer in the decoding network.
We train the model until the reconstruction loss stops decreasing.

\subsection{Mitigating Posterior Collapse}

For the structured part $\pmb z^{(1)}$, we use $\beta$-VAE setting $\beta$ as 0.2 across all the datasets.
For the unstructured part $\pmb z^{(2)}$, different strategies are employed for each dataset:

\begin{itemize} 
\item {\bf Yelp}: $\beta$-VAE setting $\beta$ as 0.35.
\item {\bf Amazon}: $\beta$-VAE setting $\beta$ as 0.35.
\item {\bf AG-News}: KL annealing, from 0.1 to 1.0 in 10 epochs.
\end{itemize}

\subsection{Hyperparameter Settings}

 \begin{table}[h]

\centering
\caption{Hyperparameter settings.}
\scriptsize
\begin{tabular}{| l | l | l | l |}
\hline
& Yelp  & Amazon & AG-News \\ \hline
Number of variations $K$ & 3 & 3 & 10 \\
Parameter to control the KL $\alpha$ & 100 & 100 & 10 \\
Input dimension for LSTM encoder & 256 & 256 & 512 \\
Hidden dimension for LSTM encoder & 1024 & 1024 & 1024 \\
Dimension for $\pmb{z}^{(2)}$ & 64 & 64 & 96 \\
Dimension for $\pmb{z}^{(1)}$ & 16 & 16 & 32\\
Input dimension for LSTM decoder & 128 & 128 & 512\\
Hidden dimension for LSTM decoder & 1024 & 1024 & 1024\\
\hline
\end{tabular}
\end{table}

We choose $K \in \{3, 5, 10\}$, $\alpha \in \{1, 10, 100\}$, input dimension for LSTM encoder $\in \{128, 256, 512\}$, hidden dimension for LSTM encoder $\in \{512, 1024, 2048\}$, dimension for $\pmb{z}^{(2)} \in \{32, 64, 96\}$, dimension for $\pmb{z}^{(1)} \in \{16, 32, 48\}$, input dimension for LSTM decoder $\in \{128, 256, 512\}$ and hidden dimension for LSTM decoder $\in \{512, 1024, 2048\}$. 
{\bf Amazon} follows the same setting as {\bf Yelp} without extra tuning. 

For hyperparameter tuning, one cannot rely on a single metric to tune the hyperparameters due to the inherent trade-off between transfer strength and content preservation.
Instead, we search the above grids to find a setting with low reconstruction loss and high KL loss.
In other words, it is likely that the setting we found is not the optimal one.
Automatic approach could use multi-objective optimization to find the Pareto optimal setting.
But we do not feel it is necessary here as our method is relatively insensitive to the hyperparameters except $\beta$ used to control the KL loss.
For $\beta$, we inspect the generated outputs on the training set to decide its value.

\subsection{Decoding Strategy}

For decoding, we use beam search with a beam size of 5.

\section{Proof of Minimalization of Eq.~\ref{eq:factor}}
\label{appendix:proof}

The problem can be formulated as an optimization problem as follows:

$$
\text{maximize} \sum_{i=1}^K p_i^2 ,\quad \text{subject to} \sum_{i=1}^K p_i = 1.
$$

By introducing a Lagrange multiplier $\lambda$, the Lagrange function is defined as 

$$\mathcal{L}(p_1, p_2, \dots, p_K, \lambda) = \sum_{i=1}^K p_i^2 - \lambda (\sum_{i=1}^K p_i - 1).$$

In order to find the optimal point, we require that 

$$
\frac{\partial}{\partial p_i}\left(\sum_{i=1}^K p_i^2 - \lambda (\sum_{i=1}^K p_i-1)\right)=2p_i - \lambda = 0, \quad i = 1, 2, \dots, K,
$$

which shows that all $p_i$ are equal. By using the constraint $\sum_ip_i=1$, we find $p_i=\frac1K, i=1,2, \dots,K$.
By plugging into the results, $\pmb\mu^\top\pmb\mu=\alpha\sum_i p_i^2$ reaches its minimum $\frac\alpha{K}$.

\section{Comparisons with Baselines on Topic Modelling}
\label{appendix:topic}

\paragraph{Experimental setup:}
We use the AG news dataset for this task constructed by \citep{zhang2015character}.
It contains four topic categories which are {\em World, Sports, Business} and {\em Sci/Tech}, with the title and description fields.
For each category, there are $30,000$ training samples and $1,900$ test samples.
In this paper, we drop the title and just use the description field.
We compare our approach to two standard baselines for unsupervised topic modelling: (1) {\bf LDA} \citep{blei2003latent}, a standard implementation of LDA is used for this baseline\footnote{https://radimrehurek.com/gensim/}; (2) {\bf $k$-means}. To show the power of our approach beyond the pre-trained sentence representations, we perform $k$-means clustering directly on the sentence representations.
Following \citep{manning2010introduction}, we assign each inferred topic to one of the gold-standard topics with the optimal mapping and report the precision ({\em a.k.a.} purity), recall ({\em a.k.a.} collocation) and $F_1$ score.
The number of topics is chosen to be $10$.
The results reported for the baselines and our model are the average over 10 runs.

\paragraph{Quantitative results:}
The results are shown in Table~\ref{tab:topics}.
We can see that our approach achieves comparable results to {\bf LDA} while significantly outperforming {\bf $k$-means} in all four categories, indicating that our approach can go beyond just clustering on pre-trained sentence representations.

\begin{table*}[h]
\centering
\footnotesize
\caption{Results for topic identification.}
\begin{tabular}{l | l | l l l}
\hline
Topic & Model & Precision & Recall & $F_1$ \\ \hline
\multirow{3}{*}{World} & LDA & 69.73 & {\bf 75.32} & 72.14 \\
& $k$-means & 67.64 & 47.63 & 55.90 \\ 
& Ours & {\bf 80.83} & 70.55 & {\bf 74.59} \\ \hline
\multirow{3}{*}{Sports} & LDA & 79.17 & 82.50 & {\bf 80.22} \\
& $k$-means & 47.66 & {\bf 89.50} & 62.04 \\
& Ours & {\bf 81.14} & 78.88 & 79.49 \\ \hline
\multirow{3}{*}{Business} & LDA & {\bf 72.10} & 66.45 & {\bf 68.46} \\
& $k$-means & 53.06 & 53.16 & 53.11 \\
& Ours & 64.04 & {\bf 64.53} & 63.97 \\ \hline
\multirow{3}{*}{Sci/Tech} & LDA & 66.55 & 59.77 & 61.60 \\
& $k$-means & {\bf 81.32} & 31.59 & 44.67 \\
& Ours & 65.20 & {\bf 71.74} & {\bf 66.77} \\
\hline
\end{tabular}
\label{tab:topics}
\end{table*}

\clearpage
\section{Text Transfer Examples}
\label{appendix:transfer}

\subsection{Sentiment manipulation on Yelp dataset}
\begin{table*}[h]
\centering
\small
\caption{Sentiment manipulation results from positive to negative}
\begin{tabular}{| l | p{11cm} |}
\hline
SRC & this was the best i have ever had !\\ \hline
B-GST &  this was the worst place i have ever had !\\ \hline
{\bf CP-G} & this was the worst pizza i have ever had !\\ \hline
{\bf CP-B} & this was the worst i have ever had !\\ \hline
\hline
SRC & friendly and welcoming with a fun atmosphere and terrific food .\\ \hline
B-GST &  the hummus is ridiculously bland and bland .\\ \hline
{\bf CP-G} & rude and unorganized with a terrible atmosphere and coffee .\\ \hline
{\bf CP-B} & the hummus is ridiculously greasy and tasteless .\\ \hline
\hline
SRC & i ordered the carne asada steak and it was cooked perfectly !\\ \hline
B-GST &  i ordered the carne asada steak and it was just as bad !\\ \hline
{\bf CP-G} & i ordered the carne asada steak and it was n't cooked and it was lacking .\\ \hline
{\bf CP-B} & i ordered the carne asada burrito and it was mediocre .\\ \hline
\hline
SRC &  the owner is a hoot and the facility is very accommodating .\\ \hline
B-GST &  the owner is a jerk and the facility is very outdated .\\ \hline
{\bf CP-G} & the owner is a hoot and the facility is empty and the layout is empty .\\ \hline
{\bf CP-B} & the owner is a riot and the facility is very clean.\\ \hline
\hline
SRC & i will be going back and enjoying this great place !\\ \hline
B-GST &  i wo n't be going back and this place is horrible !\\ \hline
{\bf CP-G} & i will be going back and eat this pizza hut elsewhere .\\ \hline
{\bf CP-B} & i will be going back and hated the worst dining experience .\\ \hline
\end{tabular}
\end{table*}

\begin{table*}[h]
\centering
\small
\caption{Sentiment manipulation results from negative to positive}
\begin{tabular}{| l | p{11cm} |}
\hline
SRC & there is definitely not enough room in that part of the venue . \\ \hline
B-GST & there is plenty enough seating in that part of the venue .\\ \hline
{\bf CP-G} & there is definitely an authentic dinner in that part .\\ \hline
{\bf CP-B} & there is definitely a nice theatre in that part .\\ \hline
\hline
SRC & but it probably sucks too ! \\ \hline
B-GST & but it tastes great too !\\ \hline
{\bf CP-G} & but it 's truly fun and insanely delicious .\\ \hline
{\bf CP-B} & but it 's probably wonderful when u !\\ \hline
\hline
SRC & always rude in their tone and always have shitty customer service ! \\ \hline
B-GST & always in tune with their tone and have great customer service .\\ \hline
{\bf CP-G} & always great with their birthdays and always excellent music .\\ \hline
{\bf CP-B} & always accommodating and my dog is always on family .\\ \hline
\hline
SRC &  i was very sick the night after . \\ \hline
B-GST & i was very happy the night after .\\ \hline
{\bf CP-G} & i was very pleased with the night .\\ \hline
{\bf CP-B} & i was very happy with the night .\\ \hline
\hline
SRC & this is a horrible venue . \\ \hline
B-GST & this is a wonderful venue .\\ \hline
{\bf CP-G} & this is a great place for celebrating friends .\\ \hline
{\bf CP-B} & this is a great place for beginners .\\ \hline
\end{tabular}
\end{table*}

\newpage
\subsection{Sentiment Manipulation on Amazon Dataset}
\begin{table*}[h]
\centering
\small
\caption{Sentiment manipulation results from positive to negative}
\begin{tabular}{| l | p{12cm} |}
\hline
SRC & most pizza wheels that i ve seen are much smaller .\\ \hline
B-GST &  most pizza dough that i ve seen are much better .\\ \hline
{\bf CP-G} & most pizza wheels that i ve seen are much more good and are much quality .\\ \hline
{\bf CP-B} & most pizza wheels that i ve seen are much better than are much better\\ \hline
\hline
SRC & however , this is an example of how rosle got it right .\\ \hline
B-GST & however , this game is an example of how rosle loves it .\\ \hline
{\bf CP-G} & however , this is an example of how toxic . . . sad . . . obviously .\\ \hline
{\bf CP-B} & however , this is an example of how cheap . similar . cheap advice . cheap advice . similar .\\ \hline
\hline
SRC & auto shut off after num\_num hours , which is a good feature .\\ \hline
B-GST & auto shuts off after num \_ num hours , which is a shame .\\ \hline
{\bf CP-G} & whipped mask off after num\_num hours , which is slimy , which is disgusting .\\ \hline
{\bf CP-B} & auto shut off after num\_num hours, which is a stupid idea , which seems to be bad .\\ \hline
\hline
SRC &  that said , the mic did pic up everything it could .\\ \hline
B-GST &  that said, the game took up everything it could .\\ \hline
{\bf CP-G} & that said, the shampoo did nt smell him well . stopped cleaning everything . ended up smelling sick\\ \hline
{\bf CP-B} & that said, the mic did not fit everything on well , let me down it weren t cleaning\\ \hline
\hline
SRC & i also prefered tha blade weight and thickness of the wustof !\\ \hline
B-GST &  i also like the blade weight and of the wustof .\\ \hline
{\bf CP-G} & i also disliked the blade weight and thickness of the materials .\\ \hline
{\bf CP-B} & i also slammed the blade weight and thickness of the wide .\\ \hline
\end{tabular}
\end{table*}

\begin{table*}[h]
\centering
\small
\caption{Sentiment manipulation results from negative to positive}
\begin{tabular}{| l | p{12cm} |}
\hline
SRC & the quality is declined quickly by heat exposure . \\ \hline
B-GST & the water is quickly drained by head exposure .\\ \hline
{\bf CP-G} & the quality is utilitarian so grinding or sandwiches .\\ \hline
{\bf CP-B} & the quality is priceless quickly by heat rises .\\ \hline
\hline
SRC & the directions were easy to follow but the quality of the easel was pathetic . \\ \hline
B-GST & the directions were easy to follow but the quality of the product was excellent .\\ \hline
{\bf CP-G} & the directions were easy to follow but the quality is good for the quality and is\\ \hline
{\bf CP-B} & the directions were easy to follow but the quality is what the quality is like the best quality of \\ \hline
\hline
SRC & multiplayer is just as bad, though thankfully not worse . \\ \hline
B-GST & quality is just as good , though thankfully not perfect .\\ \hline
{\bf CP-G} & besides it is just good , though . those . . usually . . . usually . . .\\ \hline
{\bf CP-B} & multiplayer is just as bad, though somebody s also so far not so far but no problem .\\ \hline
\hline
SRC &  another energy product that simply wastes our money . \\ \hline
B-GST & another energy product that simply saves our money .\\ \hline
{\bf CP-G} & another energy product that simply glides your pasta .\\ \hline
{\bf CP-B} & another energy product that simply wastes this money .\\ \hline
\hline
SRC & i received the wrong color and it shreds easily . \\ \hline
B-GST & i received the color and it works easily .\\ \hline
{\bf CP-G} & i low the new color and it closes easily .\\ \hline
{\bf CP-B} & i received the wrong color and it pours easily from dishwasher and dries easily on garlic easily .\\ \hline
\end{tabular}
\end{table*}

\newpage
\section{Text Transition Examples on AG news}
\label{appendix:transition}

\begin{table*}[h]
\centering
\small
\caption{Topic transition examples.}
\begin{tabular}{l | p{10cm}}
\hline
{\em World} throughout & BAGHDAD (Reuters) - Iraq 's interim prime minister , Iyad Allawi , said on Monday that the United States had no intention of withdrawing from the country to end the violence in Iraq . \\ \hline
{\em World} to {\em Sports} & BAGHDAD (Reuters) - Iraq 's interim prime minister , Iyad Allawi , said on Monday that the United States {\bf had no intention of withdrawing its troops from the country to the end of the year .}\\ \hline
{\em World} to {\em Business} & BAGHDAD (Reuters) - Iraq 's interim prime minister , Iyad Allawi , said on Monday that the United States {\bf had no intention of withdrawing its troops from the country to the country .}\\ \hline
{\em World} to {\em Sci/Tech} & BAGHDAD (Reuters) - Iraq 's interim prime minister , Iyad Allawi , said on Monday that the United States {\bf had no intention of withdrawing its uranium enrichment program to the United States .} \\ \hline \hline
{\em Sports} throughout & For the first time in four years , the US men 's basketball team won the gold medal in the men 's 400-meter medley relay . \\ \hline
{\em Sports} to {\em World} & For the first time in four years , the US men 's basketball team won the gold medal at {\bf the Athens Olympics in Athens , where the United States and the United States have agreed to a peace deal .}\\ \hline
{\em Sports} to {\em Business} & For the first time in four years , the US men 's basketball team won the gold medal at {\bf the Athens Olympics on Wednesday , with a surge in crude oil prices .}\\ \hline
{\em Sports} to {\em Sci/Tech} & For the first time in four years , the US men 's basketball team won the gold medal in {\bf the men 's Olympic basketball tournament in Beijing on Tuesday .}\\ \hline \hline
{\em Business} throughout & NEW YORK (Reuters) - U.S. stocks opened higher on Friday , as oil prices climbed above \$48 a barrel and the Federal Reserve raised interest rates by a quarter percentage point .\\ \hline
{\em Business} to {\em World} & NEW YORK (Reuters) - U.S. stocks opened higher on Friday , as oil prices climbed above \$48 a barrel {\bf and the Federal Reserve raised interest rates by a quarter percentage point .}\\ \hline
{\em Business} to {\em Sports} & NEW YORK (Reuters) - U.S. stocks opened higher on Friday , as oil prices climbed above \$48 a barrel {\bf and the Federal Reserve raised interest rates by a quarter percentage point .}\\ \hline
{\em Business} to {\em Sci/Tech} & NEW YORK (Reuters) - U.S. stocks opened higher on Friday , as oil prices climbed above \$48 a barrel {\bf and the Federal Communications Commission said it would allow the companies to use mobile phones .}\\ \hline \hline
{\em Sci/Tech} throughout & SINGAPORE (Reuters) - South Korea 's Hynix Semiconductor Inc. said on Tuesday it had developed a prototype micro fuel cell recharger for a range of security vulnerabilities in India . \\ \hline
{\em Sci/Tech} to {\em World} & SINGAPORE (Reuters) - South Korea 's Hynix Semiconductor Inc. said on Tuesday it had developed a prototype micro fuel {\bf cell aimed at ending a standoff with North Korea .}\\ \hline
{\em Sci/Tech} to {\em Sports} & SINGAPORE (Reuters) - South Korea 's Hynix Semiconductor Inc. said on Tuesday it had developed a prototype micro fuel {\bf cell aimed at protecting the world 's biggest gold medal .} \\ \hline
{\em Sci/Tech} to {\em Business} & SINGAPORE (Reuters) - South Korea 's Hynix Semiconductor Inc. said on Tuesday it had developed a prototype micro fuel {\bf cell aimed at protecting the world 's largest oil producer .} \\
\hline
\end{tabular}

\end{table*}

%
%

\end{document}